\documentclass[
  english,        
  font=times,     
  twocolumn,      
]{tumarticle}

\usepackage{lipsum}
\usepackage{tabularx}
\usepackage{longtable}
\usepackage{graphicx}
\usepackage{threeparttable}
\title{Assessing Student Errors in Experimentation Using Artificial Intelligence and Large Language Models: A Comparative Study with Human Raters}

\author[affil=1]{Arne Bewersdorff}
\author[affil=1]{Kathrin Seßler}
\author[affil=2]{Armin Baur}
\author[affil=1]{Enkelejda Kasneci}
\author[affil=1]{Claudia Nerdel}

\affil[mark=1]{\theDepartmentName, \theUniversityName}
\affil[mark=2]{University of Education Heidelberg}

\date{August 11, 2023}

\def\keywordname{{\textbf{Keywords:}}}
\providecommand{\keywords}[1]{\def\and{{\textperiodcentered} }%
\par\addvspace\baselineskip
\noindent\keywordname\enspace\ignorespaces#1}%

\begin{document}

\maketitle

\begin{abstract}
  Identifying logical errors in complex, incomplete or even contradictory and overall heterogeneous data like students’ experimentation protocols is challenging. Recognizing the limitations of current evaluation methods, we investigate the potential of Large Language Models (LLMs) for automatically identifying student errors and streamlining teacher assessments. Our aim is to provide a foundation for productive, personalized feedback. Using a dataset of 65 student protocols, an Artificial Intelligence (AI) system based on the GPT-3.5 and GPT-4 series was developed and tested against human raters. Our results indicate varying levels of accuracy in error detection between the AI system and human raters. The AI system can accurately identify many fundamental student errors, for instance, the AI system identifies when a student is focusing the hypothesis not on the dependent variable but solely on an expected observation (acc. = 0.90), when a student modifies the trials in an ongoing investigation (acc. = 1), and whether a student is conducting valid test trials (acc. = 0.82) reliably. The identification of other, usually more complex errors, like whether a student conducts a valid control trial (acc. = .60), poses a greater challenge. This research explores not only the utility of AI in educational settings, but also contributes to the understanding of the capabilities of LLMs in error detection in inquiry-based learning like experimentation.
  
\end{abstract}

\section{Introduction}
Competencies for planning and conducting scientific inquiry like experiments serve as vital components of science curricula in Germany (KMK, 2004) and around the globe (e.g. US: National Research Council, 2013; UK:  Department for Education, 2014; Finnland: Finnish National Board of Education, 2014) fostering the development of scientific thinking and problem-solving skills of students (Bewersdorff et al., 2020). Despite its importance, students frequently encounter challenges during the planning and implementation of experiments. These challenges – or errors – have been well-documented and empirically validated through numerous studies over the past two decades (Kranz et al., 2022).\\
Current tools available for the identification of these errors primarily consist of rating schemes, paper-pencil tests, or on-the-fly observations which are employed by teachers or students themselves to evaluate student performance (Hild et al., 2019; Lehtinen et al., 2022). However, these methods have several limitations. For one, they put a significant burden on teachers since they require them to meticulously review and assess each students’ work individually. Additionally, these rating schemes are often found to be complex and time-consuming, making them less accessible and user-friendly for teachers (Baur, 2015). Another critical constraint of these tools is their reliance on the reliability and objectivity of the users, primarily teachers and researchers. As a result, the feedback provided to students may not always be consistent or accurate, limiting its effectiveness in addressing common student errors and hence the improvement of experimental competencies.
In light of these limitations, there is a growing need for the development of more efficient and intuitive tools to analyze and address common errors made by students during the planning and conducting of experiments. Such tools would not only help to streamline the evaluation process for teachers but could eventually help to provide valuable feedback to students, ultimately enhancing their understanding of experimental design and fostering their development of critical scientific competencies.\\
AI models, especially Large Language Models, have emerged as a transformative force in the field of education (Abdelghani et al., 2023; Bhat et al., 2022; Dijkstra et al., 2023; Ji et al., 2023; MacNeil et al., 2023). These advanced generative AI systems like GPT-3 (Brown, 2020), ChatGPT (OpenAI, 2022), GPT-4 (OpenAI, 2023) or the recently released LaMDA model (Thoppilan et al., 2022), are deep learning architectures that use massive text datasets and reinforcement learning with human feedback to learn how to generate human-like texts. This training procedure enables them to understand and respond to a wide range of natural language queries. In educational settings, LLM-based AI systems are increasingly being leveraged for personalized learning (Murtaza et al., 2022), intelligent tutoring systems (Marmo, 2022), and supporting content generation (Khosravi et al., 2023). By being able to adapt to individual learners’ needs and providing instant feedback, LLM-based AI systems hold great potential to enhance the overall educational experience and bridge the gap between students and expert knowledge (Kasneci et al., 2023). Often LLM-based AI systems are primarily used to give rather general feedback to students. While this is a great way to support students' learning, these replies by the LLM-based AI systems are not comparable among each other and the focus of the feedback might shift leading to reduced reliability and validity. To counter this issue, we decided to aim for well-described student errors and test an LLM-based AI system for each error against humans.

\section{Framework}
\subsection{Identification of student errors during experimentation}
The first step of any student assessment is the identification of the current state of the student performance, e.g. their errors in a given task. The description of student errors during scientific inquiry, especially experimentation, has a long tradition in educational sciences, leading to a comprehensive collection of student errors (for a review see Kranz et al., 2022). While other papers employ terms like ‘problems’, ‘difficulties’, or ‘challenges’ to characterize aspects of the experimental process (e.g. Jong \& van Joolingen, 1998; Kranz et al., 2022), throughout this work and in line with Schwichow et al. (2022), but in a broader understanding, we will use the term ‘error’. The term ‘error’ in this paper refers to actions taken during the inquiry process that potentially complicate or even hinder students from reaching a conclusive result (cf. Baur, 2023).\\
To identify student errors in experimentation, the student protocols of the experiments can serve as a valid data source. The protocols provided by students shed light on the final outcome of the experiment, but offer minimal to no insights into the actual process while conducting the experiment itself. Given that student protocols only illustrate the final state and outcome of the experiment, we were compelled to exclude all errors associated with the experimental procedure itself. This limitation arises from the fact that these protocols fail to identify errors that may have occurred during the experimentation process but were not evident in the students' protocols. This led to a list of 16 common student errors, which are presented in the following Table~\ref{tab:definition_errors}.

\begin{table*}
\centering
\begin{tabular}{p{0.17\textwidth} p{0.17\textwidth} p{0.18\textwidth} p{0.18\textwidth} p{0.17\textwidth}}
 \hline
 Phase & Definition & Label & Description/Example from the sample (see 4.2) & Reference(s) \\
 \hline
 State a hypothesis & Hypothesis is not focused on the dependent variable, but on an expected observation & hyp\_var\_obs & \emph{“I think because of the water the lid pops open.”} & Baur, 2018 \\
 \hline
  & Hypothesis consists of a combination of independent variables  & hyp\_var\_comb & \emph{“I suspect that the cones contract due to the cold and the moisture.”} & Baur, 2018; Valanides et al., 2014 \\
 \hline
  & Hypothesis has no dependent variable & hyp\_no\_dep	& \emph{“It needs water.”} & - \\
 \hline
  & No hypothesis is proposed & hyp\_exists & Student works without posing a hypothesis & J. Zhai et al., 2014 \\
 \hline
 Design and conduct an experiment & Material is missing & material\_miss & The student does not itemize the material he is using & Garcia‐Mila \& Andersen, 2007 \\
 \hline
  & Missing test trial & is\_test & No trial without an independent variable & Baur, 2021\\
 \hline
  & Missing control trial & is\_control & No trial where all variables are present & Dasgupta et al., 2016; Germann et al., 1996\\
 \hline
  & Student plans and prepares experimental trials and forgets the necessary component & missing\_components & The student conducts an experiment to determine what yeast needs to produce CO2, but without using yeast & Baur, 2021\\
 \hline
  & Trials with the same content (no variation) & no\_variation & Student conducts trials with the same content and the same instruments & H.‑K. Wu \& C.‑L. Wu, 2011\\
 \hline
  & Experimental trials are altered & alter\_exp & The student (repeatedly) alters running experimental trials - they add more ingredients, remove a stopper, stir the mixture, etc. & Baur, 2021 \\
 \hline
  & Only one trial is conducted & one\_trial & The student conducts only one trial & Hammann et al., 2008 \\
 \hline
  & Documentation of the implementation is missing & no\_impl & Student does not describe his implementation & Garcia‐Mila \& Andersen, 2007\\
\end{tabular}

\end{table*}

\begin{table*}
\centering
\begin{tabular}{p{0.17\textwidth} p{0.17\textwidth} p{0.18\textwidth} p{0.18\textwidth} p{0.17\textwidth}}
 \hline
 Observe and analyze data & Observation only in one or a few trials & few\_obs & The student only observes some trials, not all, focusing primarily on one or a few & Baur, 2018 \\
 \hline
 Result \& Conclusion & Result focuses on which is the best trial, no statement about the variable(s) & best\_result & \emph{“It closes the most in water.”} & Baur, 2018 \\
 \hline
  & The students’ observation or hypotheses are given as the result & result\_obs\_hyp\_same & Student just repeats his hypotheses or observation as a result, like: “Blisters have formed.” & Boaventura et al., 2013; García-Carmona et al., 2017 \\
 \hline
  & No result & if\_no\_result & \emph{“I have no result. I think my assumption is wrong.”} & \\
 \hline
 \\
\end{tabular}
\caption{Definition and description of students’ errors eligible for identification from their experimentation protocols.}
\label{tab:definition_errors}
\end{table*}

\subsection{Current state and vision of AI in education}
Considering the tools currently available for error identification – primarily rating schemes, paper-pencil tests, and on-the-fly observations (Hild et al., 2019; Lehtinen et al., 2022) – it's evident that there is a need to develop more efficient and intuitive methods to detect common student errors during the experimentation process. AI systems have the potential to aid in this effort.\\
In general, there are strong arguments for the use of AI systems in the educational context as they could improve access to education (Osetskyi et al., 2020), foster personalized learning (Holmes et al., 2016), unlock teacher time (Sadiku et al., 2021), reduce inequality (for debate see: Holstein \& Doroudi, 2021) and therefore improve learning in general (Chen et al., 2020). Besides these promises, the use of highly integrated AI systems also comes with some potential risks. AI systems may lead to reduced student privacy (X. Zhai et al., 2021) and challenge the pedagogical relations as well as the teachers’ autonomy. Teachers and learners have a manifold of misconceptions and fears about AI (Bewersdorff et al., 2023) which might lead to general skepticism towards AI systems in the classroom among stakeholders (Douali et al., 2022), ultimately hindering its effective implementation. AI systems might even increase inequality among students (Noy \& Zhang, 2023).\\
A successful integration of AI systems in education should complement teachers on their mission to foster students’ learning. Therefore, it is crucial to understand that AI systems should not replace the teacher, or, worse, be seen as competitors, but that the AI system alters their role in the learning process (Burbules et al., 2020; Schiff, 2020). The teachers’ role changes depending on the degree of automation. Different modes and degrees of implementation of AI systems are imaginable (e.g.: Six levels of automation of AI in education: I. Teacher only, II. Teacher assistance, III. Partial automation, IV. Conditional automation, V. High automation and VI. Full automation; Molenar, 2022). This can range from using the AI system to provide supportive information (II.) to automatically controlling the entire learning process (VI.). In line with Molenaar et al. (2017), our goal is a hybrid intelligence with combined responsibility between the AI system and the teacher. For a successful integration, we argue that it is crucial to respect the teachers’ autonomy and therefore give them full freedom to decide to which degree they want to use AI systems.\\ 
The shift towards a hybrid intelligence would give teachers more time to concentrate on clarifying concepts, fostering students’ critical thinking skills, encouraging creativity, and fostering an engaging and dynamic learning environment while still being in full control of the learning process and associated pedagogical considerations.
\subsection{AI based assessment in science education}
A promising application of AI systems in education is assessing student outcomes in science education. There are two general approaches of assessment: formative assessment which is ongoing during the learning process, and summative assessment at the end of a learning unit (Harlen \& James, 1997). Formative assessment, as a practice inherent in teaching that focuses on the learning process, is intended to help continuously adapt the teaching to the needs of the students (Filsecker \& Kerres, 2012). It has been, despite some critics (Bennett, 2010), identified as one of the most significant influencing factors for effective learning (Hattie, 2009), especially forms of (computer-based) ‘rapid formative assessment’ have been shown to be highly effective (Yeh, 2010). AI driven systems could help teachers with formative assessment (Swiecki et al., 2022).\\
Some educational researchers in the field of science education raise concerns about the use of AI systems for formative assessment (Li et al., 2023). Central points of critique are the confinement of the pedagogical facet of assessment and the sidelining of professional expertise as well as that AI based assessment might only evaluate limited forms of learning and lead to a surveillance pedagogy (Swiecki et al., 2022). Other voices argue that AI systems are already being widely employed in formative assessment across various educational contexts and call for a shift in perspective, from viewing AI as a problem to be solved to recognizing its potential for assessment in education (X. Zhai \& Nehm, 2023). An example for the implementation of an AI system in science assessments is the automated text analysis which is used for scoring (Zhai et al. 2020). These AI systems are validated by comparing the computer-assigned scores to human-assigned scores (Williamson et al., 2012)\\
LLM-based AI systems in the field of science education are still at an early stage. Moore et al. (2022) used GPT-3 based models to evaluate the quality of student-generated questions in a college chemistry course. They report difficulties for the automatic evaluation, with accuracies between .32 and .4 thus demonstrating one potential way to help scale student assessment by using large language models. X. Wu et al. (2023) designed an AI system for automatic scoring in science education. They report Cohens Kappa ranging from .3 to .57 and demonstrate – while still some room for improvement – the preliminary potential of the LLM-based AI system.

\section{Objectives}
The majority of contemporary AI systems in science classrooms concentrate on categorizing students’ generated responses in scientific practices, particularly in areas like explanation and argumentation (X. Zhai et al., 2020). X. Zhai et al. (2020) conclude that studies are needed which examine procedures in complex decision-making processes. This study investigates the potential of LLM-based AI systems in supporting teachers by analyzing student errors: We investigate whether automatic error identification by an LLM-based AI system is as valid and reliable as that by science educators. 

\section{Design and methods}
\subsection{Data collection}
\begin{figure*}[htbp]
    \centering
    \subfloat{
    \fbox{\includegraphics[width=0.35\textwidth]{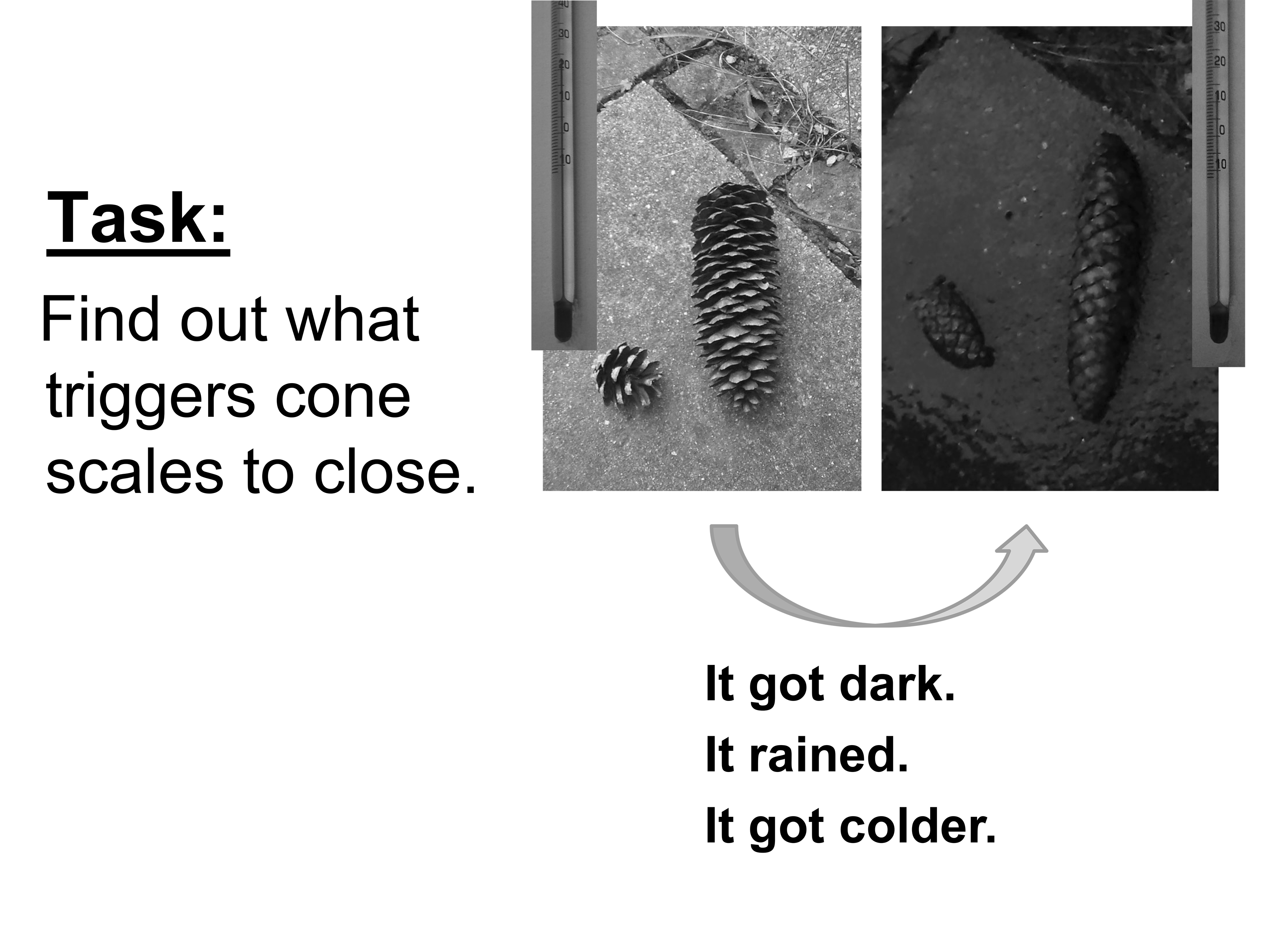}}
    }
    \subfloat{
    \fbox{\includegraphics[width=0.35\textwidth]{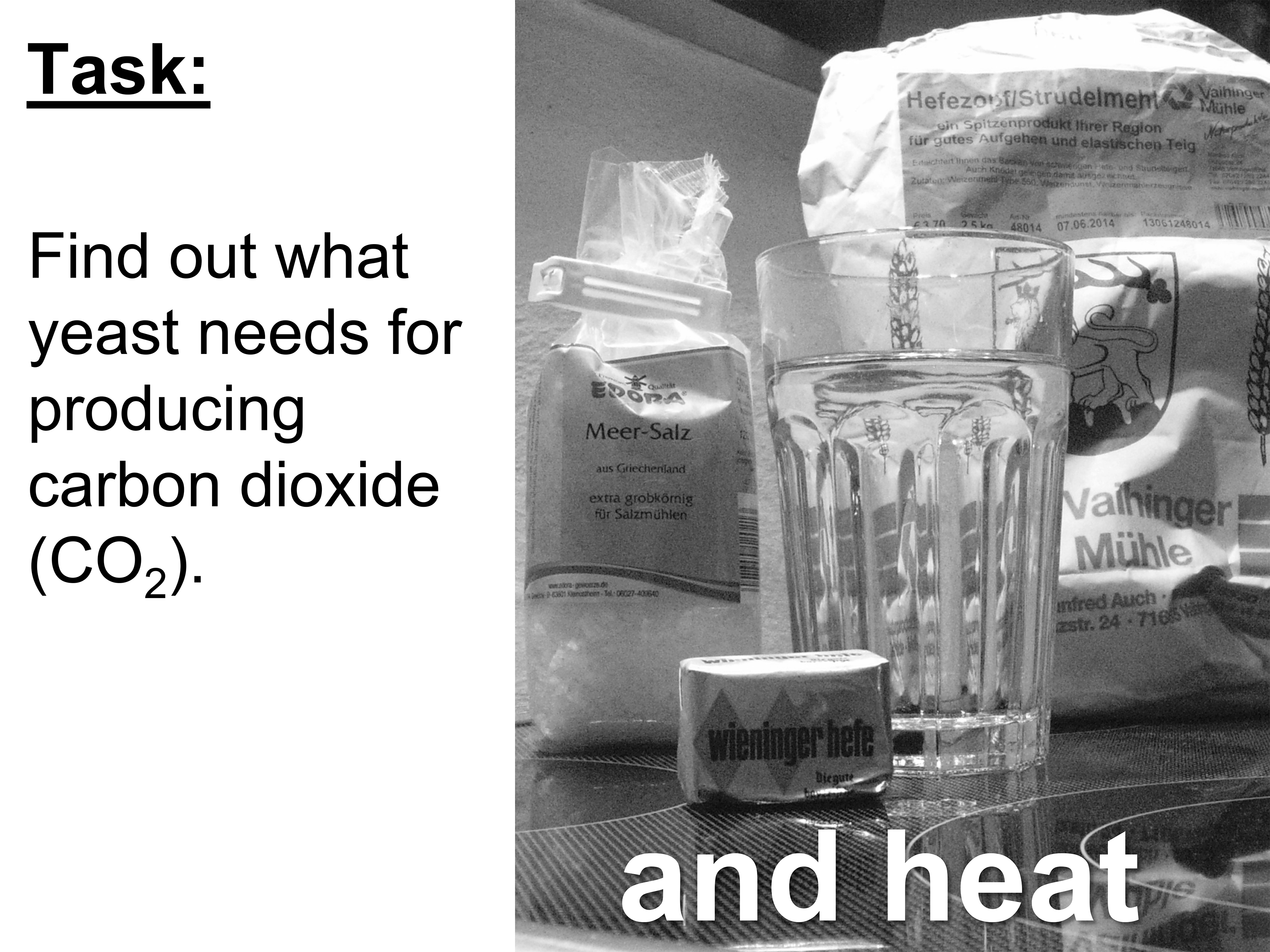}}
    }
    \caption{The two tasks given to the students to analyze their experimental procedure. The left image shows the change of a pine cone that the students were asked to reproduce. The right picture shows the material available (salt, yeast, water and flour) to stimulate  yeast to produce carbon dioxide.}
    \label{fig:tasks}
\end{figure*}

The data was gathered from a sample of 37 sixth to eighth-grade students attending secondary schools in Southern Germany. To ensure a diverse sample, the academic performance of the students was estimated by summing up their school grades in mathematics, German, and science. Teachers invited students with good, average, and poor academic performance to participate in the study. All participating students volunteered and had parental consent.\\ 
Data was collected through completing experimentation protocols with sections for ‘Hypothesis’, ‘Material’, ‘Sketch of the experimental setup’, ‘Description of the implementation’, ‘Observation’, and ‘Result’. They were given two tasks (Figure~\ref{fig:tasks}) where they had to plan, execute, and evaluate experiments. The first task involved a yeast experiment, where students had to determine the conditions necessary for yeast to produce carbon dioxide (task: “Find out what yeast needs to produce carbon dioxide”). The second task required them to explore the factors causing pine cone scales to close (task: “Find out what triggers cone scales to close”). Various materials were provided for each task, and students were free to choose which materials to use.
\noindent Both tasks were completed either on the same day or on two consecutive days, with 60 minutes allotted for each experiment. The students worked independently, supervised by a trained university student as assistant. The university student assistants were responsible for ensuring task comprehension, explaining the experimentation protocol and reminding students to continue documenting their work. They did not assist in conducting the experiments.
\subsection{Sample} 
The final dataset itself consists of 65 structured student protocols in German language from laboratory conditions, focusing on experiments related to cones and yeast. 25 protocols were rated by human raters and then exclusively used as training data for adapting and revising the AI system. The remaining 40 protocols were exclusively used for calculating inter-rater agreement between humans and the AI (all 40 protocols) and between three human raters (15 protocols as a subset of the 40 protocols). The 40 protocols for calculating the inter-rater agreement between humans and the AI system as well as the 15 protocols for calculating the inter-rater agreement between three humans respectively were selected with respect to a diverse sample regarding the topic of the experiment (cones or yeast), the students’ gender (female or male), their grade ($5^{th}$, $6^{th}$, $7^{th}$ or $8^{th}$ grade) as well as their academic performance (poor, average, good). The composition is displayed in Table~\ref{tab:irr_samples}.
\begin{table*}[t]
\centering
\begin{tabular}{p{0.20\textwidth} p{0.12\textwidth} p{0.12\textwidth} p{0.12\textwidth} p{0.12\textwidth} p{0.12\textwidth}}
 \hline
 Dataset & Topic & Gender & Grade & Academic performance & Total \\
 \hline
 training dataset & cones: 10\newline yeast: 15 & female: 13\newline male: 12 & $6^{th}$:7\newline $7^{th}$: 11\newline $8^{th}$: 7 & poor: 7\newline average: 9\newline good: 9 & 25\\
 \hline
 Inter-human rating (subset of Human vs. AI dataset) & cones: 7\newline yeast: 8 & female: 7\newline male: 8 & $6^{th}$:5\newline $7^{th}$: 5\newline $8^{th}$: 5 & poor: 4\newline average: 6\newline good: 5 & 15\\
 \hline
 Human vs. AI & cones: 20\newline yeast: 20 & female: 20\newline male: 20 & $6^{th}$:15\newline $7^{th}$: 13\newline $8^{th}$: 12 & poor: 13\newline average: 13\newline good: 14 & 40\\
 \hline
 \\
\end{tabular}
\caption{Composition of the samples for calculating the inter-rater agreement}
\label{tab:irr_samples}
\end{table*}
\subsection{Development of the AI system}
In our project, we use a pre-trained Large Language Model to analyze the experimental protocols for common student errors. Due to the inherent strength of pre-trained LLMs to follow textual descriptions (Brown, 2020), we can leverage their capabilities and operate on our limited training dataset of merely 25 student protocols. Mitigating the need for a larger, custom dataset, we can exploit their knowledge and make accurate predictions and assessments.\\
A valid and published rating scheme, encompassing common student errors but originally focusing on videotaped analysis (Baur, 2021), served as the foundation to build an AI system for detecting these errors. This AI system is based on models of the GPT-3.5 series (Ouyang et al., 2022) as well as the GPT-4 series (OpenAI, 2023), specifically using the "*-0613" snapshots corresponding to the versions of the GPT models from June 2023. \\
We used different prompting techniques. A ‘prompt’ is typically a short string of text that includes instructions for the task (zero-shot learning) or a few samples of the task (few shots learning) (Liu et al., 2022; Mayer et al., 2023). To customize the LLMs for our use case, we used Chain-of-Thought prompting (Wei et al., 2022) and role prompting (defining GPTs role as “You are a science teacher looking at student's protocols of experiments” or similar) among others. \\
Besides the myriad challenges associated with deploying AI systems based on LLMs for general feedback, a particularly prominent issue is accurately identifying logical errors in complex, incomplete or even contradictory data like students’ experimentation protocols. For this task, we generally followed a two-pronged approach. Firstly, we identified critical elements of the experiment, such as the dependent and independent variables by dissecting and understanding the students’ hypothesis. This initial step forms the basis for understanding the structure and design of the whole experiment. Secondly, we performed a systematic and algorithmic amalgamation of these identified elements. This involves examining e.g. whether the number of independent variables aligns with the number of test trials conducted. Any discrepancy may point towards errors made in the experimental process. Therefore, this complex process of error identification requires an interplay of methods, based on both LLMs and purely algorithmic procedures.
\begin{figure*}[htbp]
    \centering
    \includegraphics[width=\textwidth]{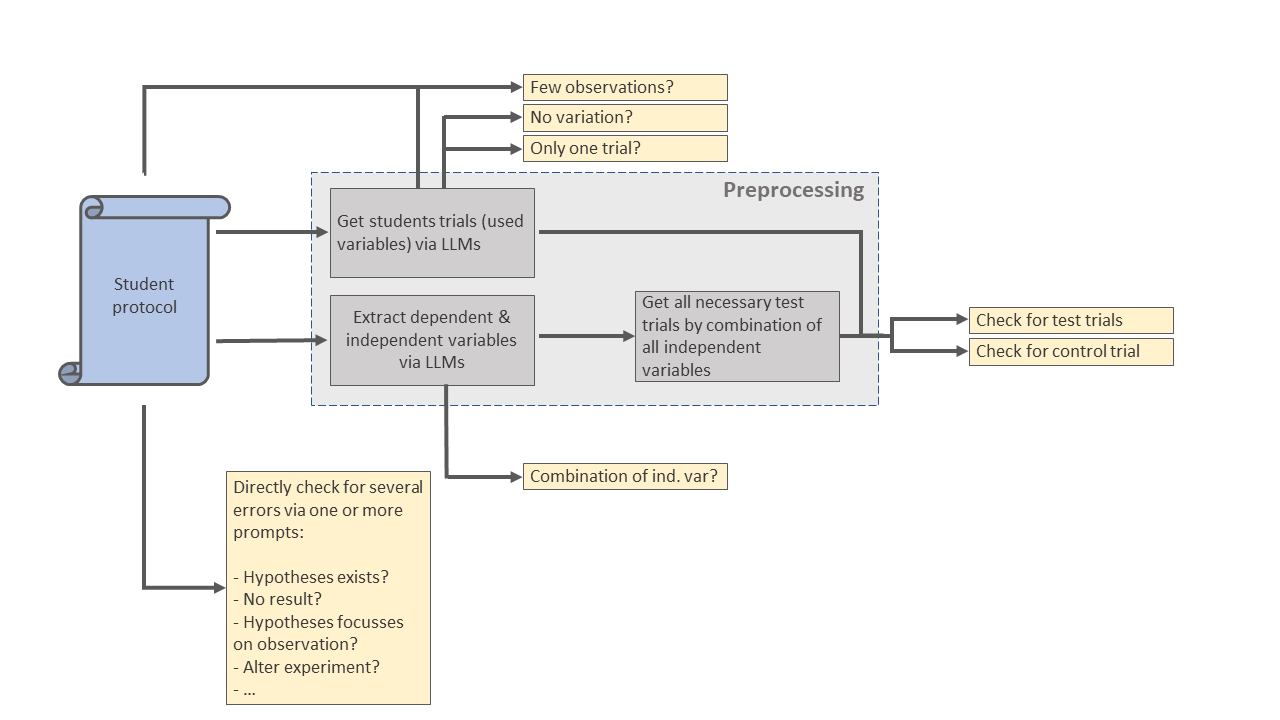}
    \caption{Simplified flow chart of our developed LLM-based AI system}
    \label{fig:flowchart}
\end{figure*}
\noindent The yellow boxes in Figure~\ref{fig:flowchart} represent the LLM-based methods for identifying the student errors. The arrows indicate the flow of information. Some errors can be directly identified through prompting and text from the students’ protocol, while others undergo preprocessing before actually checking if the error is present. In this preprocessing stage (gray boxes), key features of the experiment are extracted from the protocol using both LLMs (prompting) and algorithmic techniques. The identification of these errors is then performed based on this extracted information.
\subsection{Methods of data analysis}
The aim of the data analysis is twofold. First, our objective is to compare how different human raters, guided by the shared rating scheme, rate the student protocols. Second, it aims to compare these results with the outcomes produced when the AI system is applied on student protocols. Therefore, first we calculated inter-rater agreement among three human raters. We employed three human raters to rigorously assess the instruments’ reliability, going beyond the typical inter-rater agreement derived from two raters. The agreement between three human raters provides us an essential benchmark, demonstrating the consistency, validity and replicability of our rating scheme to detect student errors across different individuals. Next, we calculated the inter-rater agreement between human raters and the AI system. The agreement between human raters and the AI system allows us to ascertain that the AI system is capable of replicating the same process accurately and demonstrating its efficacy. As a by-product the inter-rater agreement between humans and the AI system is adding another layer of validity to the used rating scheme.\\
For the evaluation of the effectiveness of the AI system, different methods established in the field of computer science as well as methods common in the field of social sciences are applied. The ratings are compared between human raters and the AI-generated analyses by metrics common in the field of AI (Accuracy) and in the field of social sciences (Cohens Kappa, Cohen, 1960; Fleiss Kappa, Fleiss, 1971 and Gwet’s AC1, Gwet, 2014). For classification tasks in computer science, accuracy is a commonly used performance metric to evaluate the effectiveness of a model. Accuracy measures the proportion of correctly classified instances out of the total number of instances (Goodfellow et al., 2016). To calculate inter-rater reliability among two raters we use Cohens Kappa, for three raters, we use Fleiss Kappa (Fleiss, 1971). We enhance our overview of inter-rater agreement by incorporating Gwet’s AC1. Gwet’s AC1 metric provides a more stable inter-rater reliability coefficient than Cohens Kappa and is less affected by prevalence and marginal probability than Cohens Kappa (Wongpakaran et al., 2013). As we report on many errors which are very common or very rare we use Gwet’s AC1 for our inter-rater reliability analysis to gain more interpretable results.\\
For the metrics Cohens Kappa, Fleiss’ Kappa and Gwet’s AC1 the scale proposed by Landis \& Koch (1977) is applicable and used throughout this paper: .00 - .20: slight agreement; .21 - .40: fair agreement; .41 - .60: moderate agreement; .61 - .80: substantial agreement and .81 - 1: almost perfect agreement. 

\section{Results}
\subsection{Inter-rater agreement between humans}
For the calculation of the human inter-rater agreement two expert raters from the field of science education (Biology and Physics) and one computer science expert rated 15 protocols (see Table~\ref{tab:irr_agreement_human}). The results derived from the comparison of the three human raters show minimum accuracies between .71 (Error: ‘Trials with the same content’) and 1.0 (Errors: ‘No hypothesis is proposed’; ‘Material is missing’; ‘Student plans and prepares experimental trials and forgets the necessary component’; ‘Documentation of the implementation is missing’; ‘Result focuses on which is the best trial’, ‘No statement about the variable(s)’; ‘No result’). Besides the fairly high accuracies, the metrics of Fleiss Kappa and Gwet’s ACI describe a more nuanced picture: while they are high for some baseline errors (e.g. ‘No hypothesis is proposed’) and show substantial agreement for many errors, it seems like even for human raters some errors are hard to detect or agree on (e.g. ‘Only one trial is conducted’ and ‘Trials with the same content’).

\begin{table*}
\centering
\begin{tabular}{p{0.14\textwidth} p{0.1\textwidth} p{0.2\textwidth} p{0.2\textwidth} p{0.1\textwidth} p{0.1\textwidth}}
 \hline
 Description & Prevalence* & Accuracy among the raters (min(R1; R2; R3)**) & Accuracy among the raters (max(R1; R2; R3)**) & Fleiss Kappa $\kappa$ & Gwet's AC1\\
 \hline
 Hypothesis is not focused on the dependent variable, but on an expected observation & .13 & .94 & 1.00 & .78 & .95\\
 \hline
 Hypothesis consists of a combination of independent variables & .40 & .88 & 1.00 & .84 & .85\\
 \hline
 Hypothesis has no dependent variable & .13 & .94 & 1.00 & .83 & .95\\
 \hline
 No hypothesis is proposed & .13 & 1.00 & 1.00 & 1.00 & 1.00\\
 \hline
 Material is missing & .00 & 1.00 & 1.00 & $\lozenge$ & 1.00\\
 \hline
 Missing test trial & .60 & .82 & .94 & .76 & .76\\
 \hline
 Missing control trial (all variables are present) & .20 & .88 & .88 & .33 & .86\\
 \hline
 Student plans and prepares experimental trials and forgets the necessary component & .00 & 1.00 & 1.00 & $\lozenge$ & 1.00\\
 \hline
 Trials with the same content (no variation) & .20 & .71 & .88 & .26 & .73\\
 \hline
 Experimental trials are altered & .13 & .88 & .94 & .73 & .89\\
 \hline
 Only one trial is conducted & .00 & .94 & 1.00 & -.02 & .96\\
 \hline
 Documentation of the implementation is missing & .00 & 1.00 & 1.00 & $\lozenge$ & 1.00\\
 \hline
 Observation only in one or a few trials & .07 & .88 & 1.00 & .56 & .90\\
 \hline
 \\
\end{tabular}
\end{table*}

\begin{table*}[t]
\centering
\begin{threeparttable}
\begin{tabular}{p{0.14\textwidth} p{0.1\textwidth} p{0.2\textwidth} p{0.2\textwidth} p{0.1\textwidth} p{0.1\textwidth}}
 \hline
 Result focuses on which is the best trial, no statement about the variable(s) & .00 & 1.00 & 1.00 & $\lozenge$ & 1.00\\
 \hline
 The students’ observation or hypotheses are given as the result & .07 & .88 & .94 & .62 & .90\\
 \hline
 No result & .07 & 1.00 & 1.00 & 1.00 & 1.00\\
 \hline
 \\
\end{tabular}
\begin{tablenotes}
\item[*]{Prevalence was determined by calculating the percentage of the median rating given by three human raters, out of a total of 15: \emph{percentage(median (R1; R2; R3)}.}
\item[**]{R1, R2, R3 = Rater 1, 2 and 3.}
\item[$\lozenge$]{not calculable, due to division by zero. This occurs when the marginals for that category are 0 in the confusion matrix.}
\\
\end{tablenotes}
\caption{Inter-rater agreement between three human raters: Accuracy, Fleiss Kappa and Gwet’s AC1}
\label{tab:irr_agreement_human}
\end{threeparttable}

\end{table*}

\subsection{Inter-rater agreement between human raters and AI}
The results derived from comparing human and AI ratings of 40 protocols show accuracies between .38 (Error: The students’ observation or hypotheses are given as the result) and 1.0 (Errors: ‘Material is missing’; ‘Experimental trials are altered’; ‘Documentation of the implementation is missing’; ‘Observation only in one or a few trials’; ‘Result focuses on which is the best trial, no statement about the variable(s)’). Besides calculating the accuracy, we also calculated Gwet’s AC1 and Cohens Kappa. For a comprehensive overview see Table~\ref{tab:irr_agreement_ai}.\\
\begin{table*}
\centering
\begin{threeparttable}
\begin{tabular}{p{0.2\textwidth} p{0.15\textwidth} p{0.15\textwidth} p{0.15\textwidth} p{0.15\textwidth}}
 \hline
 Description & Prevalence* & Accuracy & Cohens Kappa $\kappa$ & Gwet's AC1\\
 \hline
 Hypothesis is not focused on the dependent variable, but on an expected observation & .03 & .90 & -.04 & .89\\
 \hline
 Hypothesis consists of a combination of independent variables & .63 & .90 & .80 & .80\\
 \hline
 Hypothesis has no dependent variable & .13 & .78 & .05 & .71\\
 \hline
 No hypothesis is proposed & .00 & .92 & 0 & .92\\
 \hline
 Material is missing & .00 & 1 & $\lozenge$ & 1\\
 \hline
 Missing test trial & .73 & .82 & .62 & .68\\
 \hline
 Missing control trial (all variables are present) & .38 & .60 & .02 & .36\\
 \hline
 Student plans and prepares experimental trials and forgets the necessary component & .40 & .65 & .15 & .46\\
 \hline
 Trials with the same content (no variation) & .58 & .62 & .30 & .27\\
 \hline
 Experimental trials are altered & .13 & 1 & 1 & 1\\
 \hline
 Only one trial is conducted & .23 & .82 & .31 & .77\\
 \hline
 Documentation of the implementation is missing & .00 & 1 & $\lozenge$ & 1\\
 \hline
 Observation only in one or a few trials & .13 & 1 & 1 & 1\\
 \hline
 Result focuses on which is the best trial, no statement about the variable(s) & .05 & 1 & 1 & 1\\
 \hline
 The students’ observation or hypotheses are given as the result & .73 & .38 & .08 & -.21\\
 \hline
 No result & .05 & .92 & .36 & .92\\
 \hline
 \\
\end{tabular}
\begin{tablenotes}
\item[*]{Prevalence was determined by calculating the percentage of the rating given by the AI, out of a total of 40}
\item[$\lozenge$]{not calculable, due to division by zero. This occurs when the marginals for that category are 0 in the confusion matrix.}
\\
\end{tablenotes}
\caption{Inter-rater agreement between human raters and AI: Accuracy, Cohens Kappa and Gwet’s AC1}
\label{tab:irr_agreement_ai}
\end{threeparttable}

\end{table*}

The AI system demonstrates good alignment with human raters for most errors while there are some errors that require further refinement. Apart from demonstrating relatively high accuracy and substantial, or even almost perfect reliability based on the AC1 values, the Fleiss and Cohens Kappa values frequently remain within the range of fair to slight agreement. This phenomenon predominantly arises for errors that are either extremely rare or very common (e.g.: ‘No hypothesis is proposed’).

\begin{figure*}[ht]
    \centering
    \includegraphics[width=0.75\textwidth]{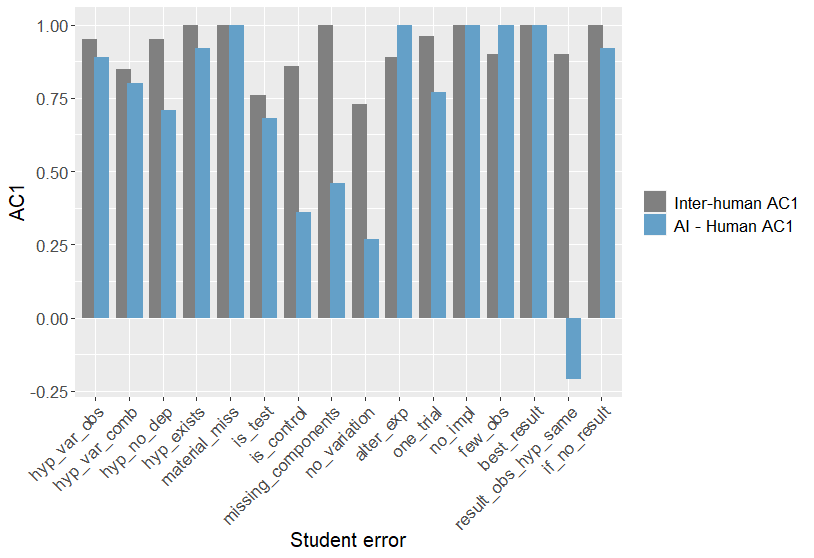}
    \caption{Comparison of AC1 values of the inter-human rating with the inter-rater reliability between raters and AI. For the errors corresponding to the labels, see Table~\ref{tab:definition_errors}}
    \label{fig:ac1_values}
\end{figure*}

\noindent Figure~\ref{fig:ac1_values} illustrates that the performance of the AI system is often quite comparable to human raters, occasionally even surpassing them (in the case of ‘Experimental trials are altered’) or matching their capabilities. However, in identifying certain errors such as ‘Missing control trial’, ‘Student plans and prepares experimental trials and forgets the necessary component’, ‘Trials with the same content’, and ‘The students’ observation or hypotheses are given as the result’, human raters still significantly outperform the AI system.

\section{Discussion}
Large language model-based AI systems, such as OpenAI's GPT series, are already applied in educational settings to generate open-ended tasks or inquiries for students to engage with (Küchemann et al., 2023), to provide a broad range of feedback (Dai et al., 2023) or as agents to think with (dos Santos, 2023). \\
By capitalizing on the unique capability of pre-trained LLMs to extract meaningful patterns from text data and to follow tangible instructions to make reliable predictions even with sparse data (Ouyang et al., 2022), we have demonstrated that an effective LLM-based student assessment does not require colossal datasets like it was necessary for traditional machine learning models. Therefore, our study emphasizes that not only the quantity, but also the quality and applicability of data are the decisive basis for powerful AI systems based on LLMs.\\
The identification of many fundamental student errors, such as verifying whether the hypothesis consists of a combination of variables (acc. = .90, $\kappa$ = .80, AC1 = .80), a student has altered trials (acc. = 1, $\kappa$ = 1, AC1 = 1) or whether the student is just focusing on the trial which worked best instead of making a statement about the variables (acc. = 1, $\kappa$ =1 , AC1 = 1), can be performed by an AI system with high accuracy. The identification of other errors yields notably lower accuracy and $\kappa$/AC1 values (see Table~\ref{tab:irr_agreement_ai} and Figure~\ref{fig:ac1_values}). This could be attributed to the frequently complex, incomplete, or even contradictory information given in student protocols. Since we attempt to identify errors purely based on the raw written results, without any form of interpretation, a reliable evaluation often falls short as younger or inexperienced students tend to provide vague descriptions of their experimental setup, which can be difficult to understand even for humans. Examples for such descriptions of their implementations are (translated from German into English): “1. Pine cone in beaker and a little water added (more water). 2. Pine cone in cardboard box. 3. Ice and cone in beaker. 4. Used the hair dryer to heat the cone. -> 1-4: All in a box 5. Cone in cooler, without the ice touching it (more ice)” and “1. Yeast mixed with water (warm), salt, and flour, filled into a test tube and a balloon placed over it. 2. The same but with cold water and cap, other mixture (cold water and more yeast), new water. 3. With the stopper, kept refilling water.”\\
One error that notably underperforms concerns the identification of whether a student has proposed a valid control trial in correlation with their suggested hypotheses (‘Missing control trial’). The identification of this error necessitates a three-step approach. First, the independent variables must be extracted from the proposed hypotheses. Second, all conducted trials have to be analyzed with respect to their inherent variables. Finally, the system must verify the existence of at least one trial that includes all independent variables identified in the hypotheses, among other variables. Complicating this process, students often slightly modify the nomenclature of the variables or fail to label them in accordance with scientific conventions. For instance, terms like “Hairdryer”, “heat”, “warm water”, “hot water”, and “water (hot)” all require an interpretation as the variable ‘heat’. This might explain to some degree why humans outperform the AI system in errors like ‘Student plans and prepares experimental trials and forgets the necessary component’ and ‘Trials with the same content’. The slight change of the nomenclature of variables or performed actions seems hard to comprehend for the AI system resulting in a lower accuracy.\\
The degree of diversity in the generated output of the LLM-based AI system is controlled via the temperature parameter. We set this parameter to zero, which results in a greedy generation process, decreasing the flexibility and variety in the output to achieve a more stable inference performance (Reiss, 2023). Still, the resulting system is not purely deterministic hindering an always reliable and reproducible assessment. Findings indicate that the behavior of LLMs, such as GPT-4, can change significantly over a relatively short period of time (Chen et al., 2023). This drift in LLMs necessitates regular monitoring of the AI system. \\
While it is generally acknowledged that bias can be a significant issue in the development of AI systems, its impact may be limited in our specific context. Our focus is the identification of errors in student protocols, a field that is less likely to encounter the sensitivities often associated with bias. Furthermore, we are optimistic that the implementation of AI systems could serve to reduce various human-induced biases (Gilovich et al., 2012), thereby enhancing the fairness of assessments. In essence, such AI systems may offer a step towards more objective and equitable educational assessments.\\
The AI system was designed as a shift towards a hybrid intelligence (Molenaar et al., 2017) supporting teachers in the assessment by identifying student errors while still leaving control over the learning process and associated pedagogical considerations with the teacher. The integration of this AI system into science classrooms paves the way for on-the-fly assessment methods that are inherently embedded within the learning process. The AI system does not draw on any additional capabilities of the student, making it a non-invasive yet potentially effective way of evaluating student errors. This approach as it seamlessly assimilates the identification of student errors within the learning process itself does not impose a separate evaluative event or procedure. Such AI systems like the presented system might even lead to a reduction of test anxiety, fostering a more authentic and supportive learning experience. 

\section{Limitations}
While our research has yielded promising results, it is essential to note its limitations to provide a balanced interpretation and guide future investigations. \\
Although the AI system is designed to identify errors across various experimental scenarios, it was tested and verified for only two given research questions: exploring the factor that leads to the closing of pine cone scales and investigating what triggers yeast to produce CO2. Therefore, the scope of the AI systems’ applicability to other experimental contexts and domains like physics remains unverified. The robustness of the AI systems’ error detection in alternative experiments is yet to be assessed. \\
The approach of providing the research questions, while useful for focusing the research, inherently bypassed an essential aspect of the scientific process – the identification and articulation of research questions. The formulation of research questions is often a challenging step for students (Cuccio-Schirripa \& Steiner, 2000). By predefining them with given tasks, we have inadvertently missed an opportunity to identify the challenges students face in generating and refining research questions. Future studies should consider incorporating stages where students are required to pose their own research questions.\\
In our study, student protocols in German language were analyzed despite our prompts for the LLMs being given in English. The linguistic mismatch introduced an additional layer of complexity, with the language translation process potentially influencing the accuracy of error identification. The AI system therefore faced the challenge of not only identifying errors in students' protocols, but also grappling with the nuances and subtleties of language translation from German to English, which could have potentially masked or altered the nature of the errors.\\
The limited amount of testing data resulted in low prevalence for some student errors (6 times < 3 out of 40) leading to low values of Cohens Kappa while the corresponding accuracy is often high. A larger test data set would facilitate a more nuanced analysis and potentially yield more robust results.\\
The protocols employed during this research were generated under laboratory conditions, wherein students operated individually under the supervision of a university student assistant. The AI system, hence, has not been tested in common science classroom settings. Thus, a comprehensive study set within a typical science classroom environment is needed to validate the AI system's effectiveness with real-world data. \\
Another limitation is the current inability of the AI system to interpret student sketches of experimental setups. These sketches often encapsulate a wealth of information and provide ample opportunities for formative assessment. Presently, our AI systems’ capabilities are text-based. Adding the ability to interpret and evaluate sketches of experimental setups could significantly augment its reliability and validity as well as its scope in general. This is likely to become a reality as LLMs such as OpenAI's GPT multimodal features, capable of processing and understanding various types of data inputs (OpenAI, 2023), become accessible for the general public.

\section{Outlook}
We were able to demonstrate the possibility to accurately identify errors in algorithm-like problem solving tasks based on complex, incomplete or even contradictory data like students’ experimentation protocols using AI systems based on LLMs like GPT-4.\\
Our findings will contribute to the growing body of research on the application of LLM-based AI systems in educational technology, particularly in the context of assisting teachers in identifying errors in students’ experiment protocols. By leveraging LLMs to analyze student errors, teachers can develop more personalized learning plans for each student, tailored to their individual needs and conceptions. This approach not only improves the overall learning experience, but also significantly reduces the time and capacity required from teachers. Additionally, our research will delve into the potential benefits and limitations of using LLM-based AI systems in educational contexts, addressing concerns regarding fairness, transparency, and ethical implications. Our study offers guidance on future research directions in this field, paving the way for innovative applications of LLM-based AI systems in various educational settings.\\
Building on the results of our investigation, we plan to develop an AI-based tool that provides students with immediate feedback on their errors in experiment protocols. This tool will not only identify errors but also offer constructive suggestions for improvement, thereby empowering students to take ownership of their learning and promoting self-reflection. Furthermore, the AI system will be designed with adaptability in mind, ensuring that it can accommodate the diverse learning styles and preferences of students across different educational contexts. By facilitating real-time feedback, our AI system aims to enhance student engagement and motivation, fostering a positive learning environment where students feel supported in their pursuit of knowledge. Moreover, we envision that this AI system will not only complement the efforts of educators but also pave the way for more efficient and effective teaching practices, ultimately transforming the landscape of education for the better. This path will be facilitated by a coordinated conversation among teachers, lecturers, and the developers of the AI system, aiming to advance the development of hybrid intelligence in education.\\

\noindent\textbf{Declaration of AI and AI-assisted technologies in the writing process}\\

\noindent During the preparation of this work the authors used ChatGPT (GPT-3.5 and GPT-4) in order to improve readability and language of single sentences as some authors are not native English speakers. After using this tool, the authors reviewed and edited the content as needed and take full responsibility for the content of the publication.\\

\noindent\textbf{References}\\

\noindent Abdelghani, R., Wang, Y.‑H., Yuan, X., Wang, T., Lucas, P., Sauzéon, H., \& Oudeyer, P.‑Y. (2023). GPT-3-driven pedagogical agents for training children's curious question-asking skills. \textit{International Journal of Artificial Intelligence in Education, 167}(3), 102887.\\

\noindent Baur, A. (2015). Inwieweit eignen sich bisherige Diagnoseverfahren des Bereichs Experimentieren für
die Schulpraxis? \textit{Zeitschrift für Didaktik der Biologie (ZDB) - Biologie Lehren und Lernen, Bd. 19} https://doi.org/10.4119/zdb-1640\\

\noindent Baur, A. (2018). Fehler, Fehlkonzepte und spezifische Vorgehensweisen von Schülerinnen und
Schülern beim Experimentieren. \textit{Zeitschrift Für Didaktik Der Naturwissenschaften, 24}(1),
115–129. https://doi.org/10.1007/s40573-018-0078-7\\

\noindent Baur, A. (2021). Errors made by 5th-, 6th-, and 9th-graders when planning and performing
experiments: Results of video-based comparisons (45-63 / Zeitschrift für Didaktik der
Biologie (ZDB) - Biologie Lehren und Lernen, Vol. 25 (2021)).\\

\noindent Baur, A. (2023). Which student problems in experimentation are related to one another? \textit{International
Journal of Science Education,} 1–25.\\

\noindent Bennett, R. E. (2010). Cognitively Based Assessment of, for, and as Learning (CBAL): A Preliminary Theory of Action for Summative and Formative Assessment. Measurement:\textit{ Interdisciplinary Research \& Perspective, 8}(2-3), 70–91. https://doi.org/10.1080/15366367.2010.508686\\

\noindent Bewersdorff, A., Baur, A., \& Emden, M. (2020). Analyse von Unterrichtskonzepten zum
Experimentieren hinsichtlich theoretisch begründeter Unterrichtsprinzipien:
Bestandsaufnahme und kriteriale Gegenüberstellung. \textit{Zeitschrift für Didaktik der Biologie,
24}(1), 108–130.\\

\noindent Bewersdorff, A., Zhai, X., Roberts, J., \& Nerdel, C. (2023). Myths, mis- and preconceptions of
artificial intelligence: A review of the literature. \textit{Computers and Education: Artificial
Intelligence, 100143}. https://doi.org/10.1016/j.caeai.2023.100143\\

\noindent Bhat, S., Nguyen, H., Moore, S., Stamper, J., Sakr, M., \& Nyberg, E (2022). Towards Automated Generation and Evaluation of Questions in Educational Domains. In \textit{Proceedings of the 15th international conference on educational data mining, International Educational Data Mining Society}, Durham, United Kingdom (pp. 701–704).\\

\noindent Boaventura, D., Faria, C., Chagas, I., \& Galvão, C. (2013). Promoting Science Outdoor Activities for Elementary School Children: Contributions from a research laboratory. \textit{International Journal of Science Education, 35}(5), 796–814.\\

\noindent Brown, T. B., Mann, B., Ryder, N., Subbiah, M., Kaplan, J., Dhariwal, P., Neelakantan, A., Shyam, P., Sastry, G., Askell, A., Agarwal, S., Herbert-Voss, A., Krueger, G., Henighan, T., Child, R., Ramesh, A., Ziegler, D. M., Wu, J., Winter, C., . . . Amodei, D. (2020, May 28). \textit{Language Models are Few-Shot Learner}s. \\

\noindent Burbules, N. C., Fan, G., \& Repp, P. (2020). Five trends of education and technology in a sustainable future. \textit{Geography and Sustainability, 1}(2), 93–97.\\

\noindent Chen, L., Chen, P., \& Lin, Z. (2020). Artificial Intelligence in Education: A Review. \textit{IEEE Access, 8}, 75264–75278.\\

\noindent Chen, L., Zaharia, M., \& Zou, J. (2023). \textit{How is ChatGPT's behavior changing over time? }http://arxiv.org/pdf/2307.09009v1 \\

\noindent Cohen, J. (1960). A Coefficient of Agreement for Nominal Scales. \textit{Educational and Psychological Measurement, 20}(1), 37–46. https://doi.org/10.1177/001316446002000104\\

\noindent Cuccio-Schirripa, S., \& Steiner, H. E. (2000). Enhancement and Analysis of Science Question Level for Middle School Students. \textit{Journal of Research in Science Teaching, 3}7(2), 210–224.\\

\noindent Dai, W., Lin, J., Jin, F., Li, T., Tsai, Y.‑S., Gasevic, D., \& Chen, G. (2023). \textit{Can Large Language Models Provide Feedback to Students? A Case Study on ChatGPT. }https://doi.org/10.35542/osf.io/hcgzj\\

\noindent Dasgupta, A. P., Anderson, T. R., \& Pelaez, N. J. (2016). Development of the Neuron Assessment for Measuring Biology Students' Use of Experimental Design Concepts and Representations. \textit{CBE Life Sciences Education, 15}(2).\\

\noindent Department for Education (2014). \textit{The national curriculum in England - Key stages 3 and 4 framework document.}\\

\noindent Dijkstra, R., Genc, Z., Kayal, S., \& Kamps, J. (2022). \textit{Reading Comprehension Quiz Generation using Generative Pre-trained Transformers.} https://e.humanities.uva.nl/publications/2022/dijk\\
\_read22.pdf \\

\noindent dos Santos, R. P. (2023). \textit{Enhancing Physics Learning with ChatGPT, Bing Chat, and Bard as Agents-to-Think-With: A Comparative Case Study. }arXiv preprint arXiv:2306.00724.\\

\noindent Douali, L., Selmaoui, S., \& Bouab, W. (2022). Artificial Intelligence in Education: Fears and Faiths. 	\textit{International Journal of Information and Education Technology, 12}(7), 650–657.\\

\noindent Filsecker, M., \& Kerres, M. (2012). \textit{Repositioning Formative Assessment from an Educational Assessment Perspective: A Response to Dunn \& Mulvenon (2009)}. https://doi.org/10.7275/xrkr-b675\\

\noindent Finnish National Board of Education. (2014). \textit{National Core Curriculum for Basic Education 2014}. Porvoon Kirjakeskus.\\

\noindent Fleiss, J. L. (1971). Measuring nominal scale agreement among many raters. \textit{Psychological Bulletin, 76}(5), 378–382. https://doi.org/10.1037/h0031619\\

\noindent García-Carmona, A., Criado, A. M., \& Cruz-Guzmán, M. (2017). Primary pre-service teachers’ skills in planning a guided scientific inquiry. \textit{Research in Science Education, 47}(5), 989–1010.\\

\noindent Garcia‐Mila, M., \& Andersen, C. (2007). Developmental Change in Notetaking during Scientific Inquiry. \textit{International Journal of Science Education, 29}(8), 1035–1058.\\

\noindent Germann, P. J., Aram, R., Odom, A. L., \& Burke, G. (1996). Student Performance on Asking Questions, Identifying Variables, and Formulating Hypotheses. \textit{School Science and Mathematics, 96}(4), 192–201. https://doi.org/10.1111/j.1949-8594.1996.tb10224.x\\

\noindent Gilovich, T., Griffin, D., \& Kahneman, D. (2012). \textit{Heuristics and Biases.} Cambridge University Press. https://doi.org/10.1017/CBO9780511808098\\

\noindent Goodfellow, I., Bengio, Y., \& Courville, A. (2016). \textit{Deep learning. Adaptive computation and machine learning.} The MIT Press. \\

\noindent Gwet, K. L [Kilem Li]. (2014). \textit{Handbook of inter-rater reliability: The definitive guide to measuring the extent of agreement among raters} (Fourth edition). Advances Analytics LLC. \\

\noindent Hammann, M., Phan, T. T. H., Ehmer, M., \& Grimm, T. (2008). Assessing pupils' skills in experimentation. \textit{Journal of Biological Education, 42}(2), 66–72. https://doi.org/10.1080/00219266.2008.9656113\\

\noindent Harlen, W., \& James, M. (1997). Assessment and Learning: differences and relationships between formative and summative assessment. \textit{Assessment in Education: Principles, Policy \& Practice, 4}(3), 365–379. https://doi.org/10.1080/0969594970040304\\

\noindent Hattie, J. (2009). \textit{Visible learning: A synthesis of over 800 meta-analyses relating to achievement} (Reprinted.). Routledge. \\

\noindent Hild, P., Gut, C., \& Brückmann, M. (2019). Validating performance assessments: measures that may help to evaluate students’ expertise in ‘doing science’. \textit{Research in Science \& Technological Education, 3}7(4), 419–445.\\

\noindent Holmes, W., \& Luckin, R. (2016). \textit{Intelligence unleashed: An argument for AI in education}. Open ideas at Pearson. Pearson; UCL Knowledge Lab.\\

\noindent Holstein, K., \& Doroudi, S. (2021, April 27). \textit{Equity and Artificial Intelligence in Education: Will "AIEd" Amplify or Alleviate Inequities in Education?} \\

\noindent Ji, H., Han, I., \& Ko, Y. (2023). A systematic review of conversational AI in language education: focusing on the collaboration with human teachers. \textit{Journal of Research on Technology in Education, 55}(1), 48–63. https://doi.org/10.1080/15391523.2022.2142873\\

\noindent Jong, T. de, \& van Joolingen, W. R. (1998). Scientific Discovery Learning with Computer Simulations of Conceptual Domains. \textit{Review of Educational Research, 68}(2), 179. https://doi.org/10.2307/1170753\\

\noindent Kasneci, E., Sessler, K., Küchemann, S., Bannert, M., Dementieva, D., Fischer, F., Gasser, U.,
Groh, G., Günnemann, S., Hüllermeier, E., Krusche, S., Kutyniok, G., Michaeli, T., Nerdel, C.,
Pfeffer, J., Poquet, O., Sailer, M., Schmidt, A., Seidel, T., . . . Kasneci, G. (2023). ChatGPT for
good? On opportunities and challenges of large language models for education. \textit{Learning and
Individual Differences, 103}, 102274.\\

\noindent Khosravi, H., Denny, P., Moore, S., \& Stamper, J. (2023). Learnersourcing in the age of AI: Student, educator and machine partnerships for content creation. Computers and Education: \textit{Artificial Intelligence, 5}, 100151. https://doi.org/10.1016/j.caeai.2023.100151\\

\noindent KMK. (2004). \textit{Beschlüsse der Kultusministerkonferenz: Bildungsstandards im Fach Biologie für den Mittleren Schulabschluss.} München. \\

\noindent Kranz, J., Baur, A., \& Möller, A. (2022). Learners’ challenges in understanding and performing
experiments: a systematic review of the literature. \textit{Studies in Science Education}, 1–47.\\

\noindent Küchemann, S., Steinert, S., Revenga, N., Schweinberger, M., Dinc, Y., Avila, K. E., \& Kuhn, J. (2023, April 20). \textit{Physics task development of prospective physics teachers using ChatGPT}. http://arxiv.org/pdf/2304.10014v1 \\

\noindent Landis, J. R., \& Koch, G. G. (1977). The Measurement of Observer Agreement for Categorical Data. \textit{Biometrics, 33}(1), 159–174. https://doi.org/10.2307/2529310\\

\noindent Lehtinen, A., Schiffl, I., Nieminen, P., \& Baumgartner-Hirscher, N. (2022). Assessment for inquiry-based learning. In A. Baur, N. Baumgartner-Hirscher, A. Lehtinen, C. Neudecker, P. Nieminen, M. Papaevripidou, S. Rohrmann, I. Schiffl, M. Schuknecht, L. Virtbauer, \& N. Xenofontos (Eds.), \textit{Differenzierung beim Inquiry-based Learning im naturwissenschaftlichen Unterricht: Ein Differenzierungstool für das Experimentieren im Sinne des Forschenden Lernen}s (1st ed., pp. 62–78). Juventa Verlag.\\

\noindent Li, T., Reigh, E., He, P., \& Adah Miller, E. (2023). Can we and should we use artificial intelligence for formative assessment in science?\textit{ Journal of Research in Science Teaching,} 21867.\\

\noindent Liu, J., Shen, D., Zhang, Y., Dolan, B., Carin, L., \& Chen, W. (2022). What Makes Good In-Context Examples for GPT-3? In E. Agirre, M. Apidianaki, \& I. Vulić (Eds.), \textit{Proceedings of Deep Learning Inside Out (DeeLIO 2022): The 3rd Workshop on Knowledge Extraction and Integration for Deep Learning Architectures} (pp. 100–114). Association for Computational Linguistics. https://doi.org/10.18653/v1/2022.deelio-1.10\\

\noindent MacNeil, S., Tran, A., Mogil, D., Bernstein, S., Ross, E., \& Huang, Z. (2022). Generating Diverse Code Explanations using the GPT-3 Large Language Model. In J. Vahrenhold, K. Fisler, M. Hauswirth, \& D. Franklin (Eds.), \textit{Proceedings of the 2022 ACM Conference on International Computing Education Research - Volume 2} (pp. 37–39). ACM.\\

\noindent Marmo, R. (2022). Artificial Intelligence in E-Learning Systems. In J. Wang (Ed.), \textit{Encyclopedia of Data Science and Machine Learning} (pp. 1531–1545). IGI Global. https://doi.org/10.4018/978-1-7998-9220-5.ch091\\

\noindent Mayer, C. W. F., Ludwig, S., \& Brandt, S. (2023). Prompt text classifications with transformer models! An exemplary introduction to prompt-based learning with large language models. \textit{Journal of Research on Technology in Education, 55}(1), 125–141.\\

\noindent Molenaar, I. (2022). Towards hybrid human‐AI learning technologies. \textit{European Journal of Education, 57}(4), 632–645.\\

\noindent Molenaar, I., Knoop-van Campen, C. A. N., \& Hasselman, F. (2017). The effects of a learning analytics empowered technology on students' arithmetic skill development. In A. Wise, P. H. Winne, G. Lynch, X. Ochoa, I. Molenaar, S. Dawson, \& M. Hatala (Eds.), \textit{Proceedings of the Seventh International Learning Analytics \& Knowledge Conference} (614–615). ACM.\\

\noindent Moore, S., Nguyen, H. A., Bier, N., Domadia, T., \& Stamper, J. (2022). Assessing the Quality of Student-Generated Short Answer Questions Using GPT-3. In I. Hilliger, P. J. Muñoz-Merino, T. de Laet, A. Ortega-Arranz, \& T. Farrell (Eds.), \textit{Lecture Notes in Computer Science. Educating for a New Future: Making Sense of Technology-Enhanced Learning Adoption} (Vol. 13450, pp. 243–257). Springer International Publishing.\\

\noindent Murtaza, M., Ahmed, Y., Shamsi, J. A., Sherwani, F., \& Usman, M. (2022). AI-Based Personalized E-Learning Systems: Issues, Challenges, and Solutions. \textit{IEEE Access, 10,} 81323–81342.\\

\noindent National Research Council. (2013).\textit{ Next Generation Science Standards}. National Academies Press. https://doi.org/10.17226/18290 \\

\noindent Noy, S., \& Zhang, W. (2023). Experimental Evidence on the Productivity Effects of Generative Artificial Intelligence.\textit{ SSRN Electronic Journal}.\\

\noindent OpenAI. (2023). \textit{ChatGPT.} https://openai.com/\\chatgpt.\\

\noindent OpenAI. (2023). \textit{GPT-4 Technical Report}. http://\\arxiv.org/pdf/2303.08774v3\\

\noindent Osetskyi, V., Vitrenko, A., Tatomyr, I., Bilan, S., \& Hirnyk, Y. (2020). Artificial Intelligence Application In Education: Financial Implications And Prospects. \textit{Financial and Credit Activity Problems of Theory and Practice, 2}(33), 574–584.\\

\noindent Ouyang, L., Wu, J., Jiang, X., Almeida, D., Wainwright, C. L., Mishkin, P., Zhang, C., Agarwal, S., Slama, K., Ray, A., Schulman, J., Hilton, J., Kelton, F., Miller, L., Simens, M., Askell, A., Welinder, P., Christiano, P., Leike, J., \& Lowe, R. (2022). Training language models to follow instructions with human feedback. \textit{Advances in Neural Information Processing Systems, 35}, 27730-27744. https://doi.org/10.48550/arXiv.2203.02155\\

\noindent Reiss, M. V. (2023). \textit{Testing the reliability of chatgpt for text annotation and classification: A cautionary remark. }arXiv preprint. arXiv:2304.11085 \\

\noindent Sadiku, M., Ashaolu, T., Ajayi-Majebi, A., \& Musa, S. (2021). Artificial Intelligence in Education. \textit{International Journal of Scientific Advances, 2}(1), 5–13.\\

\noindent Schiff, D. (2020). Out of the laboratory and into the classroom: The future of artificial intelligence in education. \textit{AI \& Society}, 1–18. https://doi.org/10.1007/s00146-020-01033-8\\

\noindent Schwichow, M., Brandenburger, M., \& Wilbers, J. (2022). Analysis of experimental design errors in elementary school: how do students identify, interpret, and justify controlled and confounded experiments? \textit{International Journal of Science Education, 44(}1), 91–114. https://doi.org/10.1080/09500693.2021.2015544\\

\noindent Swiecki, Z., Khosravi, H., Chen, G., Martinez-Maldonado, R., Lodge, J. M., Milligan, S., Selwyn, N., \& Gašević, D. (2022). Assessment in the age of artificial intelligence. Computers and Education: \textit{Artificial Intelligence, 3}, 100075.\\

\noindent Thoppilan, R., Freitas, D. D., Hall, J., Shazeer, N., Kulshreshtha, A., Cheng, H.‑T., Jin, A., Bos, T., Baker, L., Du Yu, Li, Y., Lee, H., Zheng, H. S., Ghafouri, A., Menegali, M., Huang, Y., Krikun, M., Lepikhin, D., Qin, J., . . . Le Quoc. (2022, January 20). \textit{LaMDA: Language Models for Dialog Applications}.\\

\noindent Valanides, N., Papageorgiou, M., \& Angeli, C. (2014). Scientific Investigations of Elementary School Children. \textit{Journal of Science Education and Technology, 23}(1), 26–36.\\

\noindent Wei, J., Wang, X., Schuurmans, D., Bosma, M., Ichter, B., Xia, F., Chi, E., Le Quoc, \& Zhou, D. (2022, January 28). \textit{Chain-of-Thought Prompting Elicits Reasoning in Large Language Models}. http://arxiv.org/pdf/2201.11903v6 \\

\noindent Williamson, D. M., Xi, X., \& Breyer, F. J. (2012). A Framework for Evaluation and Use of Automated Scoring. \textit{Educational Measurement: Issues and Practice, 31}(1), 2–13. https://doi.org/10.1111/j.1745-3992.2011.00223.x\\

\noindent Wongpakaran, N., Wongpakaran, T., Wedding, D., \& Gwet, K. L [Kilem L.] (2013). A comparison of Cohen's Kappa and Gwet's AC1 when calculating inter-rater reliability coefficients: A study conducted with personality disorder samples. \textit{BMC Medical Research Methodology, 13}, 61. https://doi.org/10.1186/1471-2288-13-61\\

\noindent Wu, H.‑K., \& Wu, C.‑L. (2011). Exploring the Development of Fifth Graders’ Practical Epistemologies and Explanation Skills in Inquiry-Based Learning Classrooms. \textit{Research in Science Education, 41}(3), 319–340.\\

\noindent Wu, X., He, X., Liu, T., Liu, N., \& Zhai, X. (2023). Matching Exemplar as Next Sentence Prediction (MeNSP): Zero-Shot Prompt Learning for Automatic Scoring in Science Education. In N. Wang, G. Rebolledo-Mendez, N. Matsuda, O. C. Santos, \& V. Dimitrova (Eds.), Lecture Notes in Computer Science. \textit{Artificial Intelligence in Education} (Vol. 13916, pp. 401–413). Springer Nature Switzerland.\\

\noindent Yeh, S. S. (2010). Understanding and addressing the achievement gap through individualized instruction and formative assessment. \textit{Assessment in Education: Principles, Policy \& Practice, 17}(2), 169–182. https://doi.org/10.1080/09695941003694466\\

\noindent Zhai, J., Jocz, J. A., \& Tan, A.‑L. (2014). ‘Am I Like a Scientist?’: Primary children's images of doing science in school. \textit{International Journal of Science Education, 36}(4), 553–576. https://doi.org/10.1080/09500693.2013.791958\\

\noindent Zhai, X., Chu, X., Chai, C. S., Jong, M. S. Y., Istenic, A., Spector, M., Liu, J.‑B., Yuan, J., \& Li, Y. (2021). A Review of Artificial Intelligence (AI) in Education from 2010 to 2020. \textit{Complexity, 2021}, 1–18. https://doi.org/10.1155/2021/8812542\\

\noindent Zhai, X., \& Nehm, R. H. (2023). AI and Formative Assessment: The Train Has Left the Station.\textit{ Journal of Research in Science Teaching.}\\

\noindent Zhai, X., Yin, Y., Pellegrino, J. W., Haudek, K. C., \& Shi, L. (2020). Applying machine learning in science assessment: a systematic review. \textit{Studies in Science Education, 56}(1), 111–151.\\

\end{document}